# Bitcoin Transaction Strategy Construction Based on Deep Reinforcement Learning

Fengrui Liu[a], Yang Li[a, *], Baitong Li[a, *], Jiaxin Li[b], Huiyang Xie[c]

[a] *School of Electrical Engineering, Northeast Electric Power University, Jilin 132012, China*

[b]*School of Economics and Management, Northeast Electric Power University, Jilin 132012, China*

[c]*School of Electrical and Electronic Engineering, Hanyang University, Ansan 15588, Gyeonggi, Korea*

**ABSTRACT:** The emerging cryptocurrency market has lately received great attention for asset allocation due to its decentralization uniqueness. However, its volatility and brand new trading mode has made it challenging to devising an acceptable automatically-generating strategy. This study proposes a framework for automatic high-frequency bitcoin transactions based on a deep reinforcement learning algorithm — proximal policy optimization (PPO). The framework creatively regards the transaction process as actions, returns as awards and prices as states to align with the idea of reinforcement learning. It compares advanced machine learning-based models for static price predictions including support vector machine (SVM), multi-layer perceptron (MLP), long short-term memory (LSTM), temporal convolutional network (TCN), and Transformer by applying them to the real-time bitcoin price and the experimental results demonstrate that LSTM outperforms. Then an automatically-generating transaction strategy is constructed building on PPO with LSTM as the basis to construct the policy. Extensive empirical studies validate that the proposed method perform superiorly to various common trading strategy benchmarks for a single financial product. The approach is able to trade bitcoins in a simulated environment with synchronous data and obtains a 31.67% more return than that of the best benchmark, improving the benchmark by 12.75%. The proposed framework can earn excess returns through both the period of volatility and surge, which opens the door to research on building a single cryptocurrency trading strategy based on deep learning. Visualizations of trading the process show how the model handles high-frequency transactions to provide inspiration and demonstrate that it can be expanded to other financial products.

**1. Introduction**

Cryptocurrency is a rapidly growing asset, which was born in 2009 and gradually came into the public view after 2017, with a total market capitalization of over U.S. $1,540,000,000,000 on March 1st, 2021. It utilizes a decentralized technology called blockchain to get rid of the control of corporate entities like traditional currency. Nowadays, there are over 3000 kinds of cryptocurrency and more and more large technology enterprises and investment companies take it as an important asset allocation component.

Bitcoin is the dominant kind of cryptocurrency with a market capitalization of over U.S. $969,600,000,000, which leverages blockchain technology to form a wildly-circulated digital currency. Recently, the application scenarios of virtual currency have ushered in explosive growth, gradually forming a burgeoning investment market [1]. Thus, predicting the bitcoin time series has been commonly regarded as an important research topic.

Over the past decades, most research in bitcoin price prediction has emphasized the use of classic methods borrowed from the financial domain. Katsiampa and Paraskevi [2] explore the goodness of fit of the GARCH model in the sample of bitcoin prices but they do not make predictions out of the sample, causing difficulty in proving the model' generalization. Reference [3] forecasts the bitcoin price by utilizing the auto-regressive integrated moving average model (ARIMA), which outperforms in the sub-cycle when fluctuation is within a narrow range. But it performs significantly worse in long-term predictions or with violent fluctuations. For volatile financial products such as bitcoin, ARIMA

---

[*] Corresponding author.

E-mail address: liyangnedu@gmail.com (Yang Li), cheryllee626@gmail.com (Baitong Li).

achieves a large offset from the real data. Dian employs the α-sutte factor and obtains better performances [4], whose credibility is reduced though when it is proved that bitcoin returns could not be predicted by explanatory variables [5], leading the factor construction methods to fell into a deadlock temporarily. However, the above methods have obvious shortcomings in the process of predicting bitcoin. (1) These methods either fit the bitcoin price sequence by constructing a simple nonlinear function or focus on short-term dependence rather than the long-term trend, so the accuracy of prediction results needs to be improved. (2) These methods have been proved to be effective in the stock market, but the bitcoin market lacks intuitive fundamentals and significant related factors. Therefore, simply transforming stock market research tools to study the bitcoin market may not be as effective as expected.

Prior to the work of Jing-Zhi Huang [6], the role of pure data-driven analysis in bitcoin prediction was not taken seriously, which indicates that bitcoin price can be predicted by analyzing technical indexes and big data, with little effect from fundamentals, providing a theoretical basis for machine-learning-based predictions. Since neural networks develop, it is possible to construct complex nonlinear functions and capture the long-term dependence of sequences. Neural networks have been boosting bitcoin predictions. References [7] and [8] both build an artificial neural network (ANN) to predict the price of bitcoin, but [8] concentrates on ensemble algorithms for direction prediction, rather than price prediction, which can not give references for high-frequency trading directly. Based on [7], [9] applies the neural network auto-regression (NNAR) to complete the-next-day prediction and finds that NNAR is inferior to ARIMA in daytime prediction, demonstrating that naive neural networks are possibly not useful than traditional methods. With the development of Recurrent Neural Networks (RNNs), the long sequence prediction method has been developed unprecedentedly. LSTM provides a more credible method for long-time prediction. S. McNally, J. Roche, and S. Caton [10] compare the capability of capturing longer range dependencies between LSTM and SVR and proves LSTM is more suitable for time-series prediction. Deep learning algorithms have been widely exploited to explore the law of tendency in bitcoin price. Nevertheless, there is room for improvement of the above-mentioned studies: (1) They adapt neural networks popular in the early stage, instead of comparing cutting-edge models such as TCN and Transformer. The prediction results could be probably significantly enhanced by introducing state-of-the-art structures. (2) Much of the previous research centers around static models, which can only advise in the future trend instead of direct decision-making. Therefore, the strategy construction highly relies on the manual effort of experts, precluding their potential application in sensible investment.

It is interesting to note that another idea is to introduce other factors, especially public sentiment, to jointly predict the price of bitcoin. Various studies have assessed the efficacy of sentiment in cryptocurrency investment. Matta, M., Lunesu, I., and Marchesi, M. [11] introduce Google Trends and Twitter sentiment as supplement data and the observation conclusion is that tweets anticipate by three to four days the bitcoin trend. Cavalli, S. and Amoretti, M. [12] collect data from various sources including social media, transaction history, and financial indicators. Then it applies One-Dimensional Convolutional Neural Network (1D-CNN) to do the multivariate data analysis, which achieves 74.2% test accuracy in forecasting the direction (up and down). But studies in this field face ineluctable challenges: (1) The reliability of the correlation between introduced factors and the bitcoin price is not convincing enough. (2) Data from different sources can cause large amounts of data loss and may include lots of noise, which pose an important challenge to data cleaning and raise concerns of its application in learning price patterns.

Overall, above-mentioned studies highlight the need for bitcoin prediction, but few of them propose an effective approach of constructing a dynamic strategy of bitcoin investment. Bitcoin price fluctuates dramatically, bringing challenges to static trading strategies [13]. With the advent of state-of-the-art techniques such as reinforcement learning, it is expected to take advantage of new-coming technologies to address the challenges posed by the emerging market. Hence, this paper contributes to the literature in several aspects.

In this study, a novel framework of automatically generating high-frequency transaction strategies is constructed. The main contributions involve three aspects: (1) Advanced deep learning algorithms like LSTM, TCN, and Transformer are employed to predict the price of bitcoin according to the static data. This is the first time that, as far as we know, the most advanced deep learning techniques are compared in parallel for bitcoin prediction. (2) After comparing these models by the back-test results, LSTM is selected to build a deep reinforcement learning agent for automatic high-frequency transactions based on the PPO algorithm. Our method extends deep reinforcement learning from the application of action games to the field of financial product investment decision, where the trading actions are creatively regarded as the movement of characters and returns are regarded as scores in a game. Experimental results prove that the automatically constructed policy can receive excess returns. (3) The proposed model demonstrate the possibility of constructing a single asset high-frequency trading strategy based on limited price history information and provides a direction for the realization of automatically trading. This endows the model to have a unique value in dealing with the investment of cryptocurrency market, where bitcoin has a dominant position, the configurable assets are very limited. The transaction process generated by the agent will also provide more enlightenment for professionals.

## 2. Methodology Background

### 2.1 Long-Short Term Memory (LSTM)

LSTM [14] plays an essential role in the family of RNNs, utilizing continuous observations to learn temporal dependencies to predict the future trend. Each LSTM is a set of units that capture the data flow. These units connect from one module to another, transmit past data, and collect current data. These gates are based on the neural network layer of the sigmoid function, which enables these cells to selectively allow data to pass through or process data. Fig. 1 displays the inner scheme of LSTM.

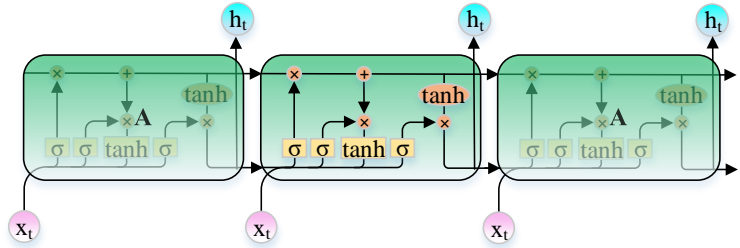

Fig. 1. LSTM

LSTM calculates the mapping from the input sequence $x = (x_1, x_2, ..., x_T)$ to the output sequence $y = (y_1, y_2, ..., y_T)$ through iterating the unit activation from $t = 1$ to $t = T$. Equations (1) - (3) denotes the calculating process of input gate, forgetting gate, and output gate respectively. Equations (4) and (5) reveal how to gain the output value of the cell at the time $t$.

$$i_t = \sigma\left(W_{ix}x_t + W_{im}m_{t-1} + W_{ic}c_{t-1} + b_i\right) \tag{1}$$

$$f_t = \sigma\left(W_{fx}x_t + W_{fm}m_{t-1} + W_{fc}c_{t-1} + b_f\right) \tag{2}$$

$$c_t = f_t \odot c_{t-1} + i_t \odot g\left(W_{cx}x_t + W_{cm}m_{t-1} + b_c\right) \tag{3}$$

$$o_t = \sigma\left(W_{ox}x_t + W_{om}m_{t-1} + W_{oc}c_t + b_o\right) \tag{4}$$

$$m_t = o_t \odot h(c_t) \tag{5}$$

where $W$ denotes the weight matrix; $m$ is the value of memory cell; $\sigma$ is the sigmoid function; $i, f$, and $o$ are the input gate, forgetting gate, output gate respectively; $b$ is the offset vector and $c$ is the unit activation vector; $\odot$ denotes the product of element direction of vector; $g$ and $h$ are the activation functions of unit input and unit output, and usually they are taken as tanh function.

### 2.2 Proximal Policy Optimization (PPO)

PPO belonging to the policy gradient (PG) method family is newly proposed by [15], which calculates the estimated value of the policy gradient and inserts it into the random gradient rising algorithm. The principle of the policy gradient method is to calculate the estimated value of the policy gradient and insert it into the random gradient rising algorithm. The most popular gradient estimator is as follows:

$$\hat{g} = \hat{E}_t \left[ \nabla_\theta \log \pi_\theta (a_t | s_t) \hat{A}_t \right] \quad (6)$$

where $\pi_\theta$ is a fixed policy; $\hat{E}_t[\cdot]$ represents the empirical average value of a limited batch of samples; $a$ denotes the action and $s$ denotes the state at time $t$; $\hat{A}_t$ is an estimator of the dominant function. The estimated value of $\hat{g}$ is got by differentiation of the objective function, which can be derived as (7):

$$L^{PG}(\theta) = \hat{E}_t \left[ \log \pi_\theta (a_t | s_t) \hat{A}_t \right] \quad (7)$$

Although using the same trajectory for multi-step optimization of lost $L^{PG}$ can achieve a better policy, it often leads to a destructive large-scale policy update, that is, the replacement policy of each step is a much too drastic improvement from the previous one, thus, it is more likely to achieve local optimization in a short time and stop iteration, unable to obtain the global optimal policy.

Based on the PG algorithm, J. Schulman et al. [16] proposes an algorithm called the Trust Region Policy Optimization (TRPO) with a creative objective function and corresponding constraints shown in (8) and (9):

$$\max_\theta \hat{E}_t \left[ \frac{\pi_\theta (a_t | s_t)}{\pi_{\theta_{old}} (a_t | s_t)} \hat{A}_t \right] \quad (8)$$

$$\hat{E}_t \left[ KL \left[ \pi_{\theta_{old}} (\cdot | s_t), \pi_\theta (\cdot | s_t) \right] \right] \leq \delta \quad (9)$$

where $\theta_{old}$ is the policy parameter vector before the update; $KL[\cdot]$ represents KL divergence. The constraint term indicates that the expected divergence of the new and old policies must be less than a certain value, which is used to constrain the change degree of each policy update. Have obtained the quadratic approximation of the constraints and the linear approximation of the target, the conjugate gradient algorithm can address the "dramatic improvement" issue effectively.

TRPO adopts constraint terms on the surface, but in fact, they are penalty terms. The above equation can be transformed into solving unrestricted optimization problems for some coefficients $\beta$, namely equation (10):

$$\max_\theta \hat{E}_t \left[ \frac{\pi_\theta (a_t | s_t)}{\pi_{\theta_{old}} (a_t | s_t)} \hat{A}_t - \beta KL \left[ \pi_{\theta_{old}} (\cdot | s_t), \pi_\theta (\cdot | s_t) \right] \right] \quad (10)$$

This is because an alternative goal forms the lower limit of the performance of policy $\pi$. TRPO has a hard constraint instead of a penalty term, because finding a proper $\beta$ value in various scenarios is especially challenging. Even in a certain scenario, different characteristics vary with the learning process. Therefore, simply setting a fixed parameter is difficult to solve the optimization problem described by the above equation. In a word, the TRPO algorithm has advantages in dealing with the task of action selection in a continuous state space, but it is sensitive to step size, so it brings insurmountable obstacles to select the appropriate step size in practical operation.

S. Kakade and J. Langford [17] modifies TRPO by proposing a novel objective function based on the method of editing agent objective. The detailed inference process is as follows:

$$r_t(\theta) = \frac{\pi_\theta (a_t | s_t)}{\pi_{\theta_{old}} (a_t | s_t)} \quad (11)$$

where $r_t(\theta)$ represents the probability ratio which is defined in (11), obviously, $r_t(\theta_{old}) = 1$. If the constraints of TRPO are removed, maximizing the original objective function will result in a policy update with too drastic changes. Therefore, it is necessary to add a penalty to avoid $r_t(\theta)$ far away from 1.

Based on the above analysis, the following objective function can be obtained as (12):

$$L^{CLIP}(\theta) = \hat{E}_t\left[\min\{r_t(\theta)\hat{A}_t, clip(r_t(\theta), 1-\varepsilon, 1+\varepsilon)\hat{A}_t\}\right] \quad (12)$$

where $\epsilon$ is a hyper-parameter, generally set to 0.1 or 0.2. The second term $clip(x_1, x_2, x_3)$ represents $max(min(x_1, x_3), x_2)$. By modifying the ratio of clipping probability to replace the target, the possibility that $r_t(\theta)$ falls outside the range [1- $\epsilon$, 1 + $\epsilon$] could be eliminated and the minimum value of cropped and un-cropped targets is taken. Hence, the lower limit of un-cropped targets becomes the ultimate goal, that is, the pessimistic bound.

The substitution loss mentioned above can be calculated and distinguished by a small change to a typical policy gradient. In practice, to realize automatic differentiation, the only necessary step is to build $L^{CLIP}$ to replace $L^{PG}$ and perform a multi-step random gradient rise on this objective.

The method of sharing parameters between the value function and the policy function has been proved to endow better performance, which requires utilizing a special architecture neural network, where a loss function combining policy substitution and error term of the value function. This purpose can be further enhanced by enlarging entropy rewards to allow ample opportunities of exploring the policy space and prevent the agent from satisfying a not-perfect-enough but acceptable action. Thus, the PPO algorithm [15] modifies the objective function as shown in (13):

$$L_t^{CLIP+VF+S}(\theta) = \hat{E}\left[L_t^{CLIP}(\theta) - c_1 L_t^{VF}(\theta) + c_2 S[\pi_\theta](s_t)\right] \quad (13)$$

where $c_1$ and $c_2$ represent parameters; $S$ denotes entropy excitation; $L_t^{VF}$ represents variance loss. J. Schulman et al. [18] proposes a policy gradient implementation method suitable for RNN. It first runs the policy of $t$ time steps, where $t$ is far smaller than episode's length, and then updates the learning strategy through employing the collected samples. An advantage estimator that looks within $T$ time steps is required as (14) shows:

$$\hat{A}_t = -V(s_t) + r_t + \gamma r_{t+1} + \ldots\ldots + \gamma^{T-t+1} r_{T-1} + \gamma^{T-t} V(s_T) \quad (14)$$

where $t$ is a certain time point in the range of $[0,T]$; $\gamma$ is the incentive discount rate in the time series. Generalized advantage estimation standardizes the above equation. Given $\lambda = 1$, it can be rewritten as (15):

$$\hat{A}_t = \delta_t + (\gamma\lambda)\delta_{t+1} + \ldots + (\gamma\lambda)^{T-t+1}\delta_{T-1}$$
$$where \quad \delta_t = r_t + \gamma V(s_{t+1}) - V(s_t) \quad (15)$$

## 3. The Proposed Approach

*3.1 Policy Function*

Deep reinforcement learning (DLP) is a combination of deep learning and reinforcement learning that integrates deep learning's strong understanding of perceived issues such as vision and natural language processing, as well as enhances decision-making capabilities for end-to-end learning.

Early reinforcement learning methods such as Q-learning [19] can only be applied to limited states and actions, which need to be designed manually in advance. However, in this scenario, the price of bitcoin can produce massive states on a long time scale. One solution is to extract features from high-dimensional data as states, and then build a reinforcement learning model. However, this approach largely depends on the design of artificial features, and in the process of dimension reduction, information of sequential dependencies will lose. Another idea is to treat bitcoin price as a continuous time-series and use a function to fit the series to form the policy. Thus, machine learning models can play the role of constructing the policy function in reinforcement learning.

This study compares traditional machine learning algorithms, neural networks, and advanced deep learning algorithms, including SVM, MLP, LSTM, TCN, and Transformer. The specific structure of these models will be introduced in the session of experiments. From the following experimental results, LSTM can best fit the historical price of bitcoin and predict the-next-day closing price, so LSTM is chosen to construct the policy of this paper.

*3.2 Reward Function*

Reward function quantifies the instant reward of a certain action and is the only information available in the interaction with the environment. Omega Ratio is selected as the reward signal, which is a performance measurement index proposed by [20], considering weighting returns and evaluating risks simultaneously, whose definition is shown in (16).

$$\omega \overset{\Delta}{=} \frac{\int_r^{\infty}(1-F(x))dx}{\int_{-\infty}^{r}F(x)dx} \quad (16)$$

where *r* is the target return threshold and *F(x)* is the cumulative distribution function of the returns.

*3.3 Bayesian Optimization*

Bayesian optimization is a technique for effectively searching the hyperspace to discover the best hyper-parameter combination to optimize the given objective function. It assumes the candidate space as compact or discrete and thus transforms the parameter-tuning problem into a sequential decision-making problem. As the iteration progresses, the algorithm continuously observes the relationship between the parameter combination and the objective function value. It selects the optimal parameter combination for the observation aim through optimizing the acquisition function, which balances the unexplored points and the best value of explored points. It also introduces regret bound to achieve state-of-art effects.

This study utilizes the Optuna tool library for Bayesian optimization. It works by modeling the objective function to be optimized using the proxy function or the distribution of the proxy function.

*3.4 Visualization*

The results are visualized to display the trading process on test data by the trained agents. The user-friendly interface is shown in Fig. 2, which is dynamic while trading. Traders can know the price of bitcoin, the actions of agents, and the corresponding net worth in real-time through the visual interface. Therefore, experts can leverage professional financial knowledge to evaluate the actions of the agent all the while obtaining enlightenment of constructing strategies from the automated trading behavior.

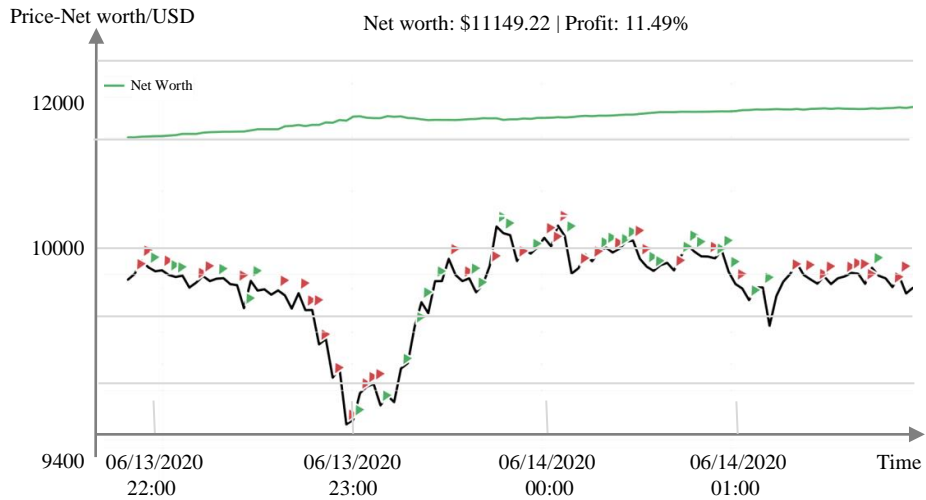

Fig. 2. Interaction Interface for the Trading Agent

In Fig. 2, the green dot indicates the point of buying while the red dot indicates the point of selling; the black line indicates the trend of bitcoin price; the green line indicates the trend of net worth.

*3.5 Scheme of Proposed framework*

Fig. 3 shows the whole process of the proposed framework. The specific experimental process is as follows:

- Create and initialize a gym trading environment.
- Setup the framework and trading sessions.
- Decide the basis of the policy function, the award function and the optimization method.
- Train and test an agent and visualize the trading process.

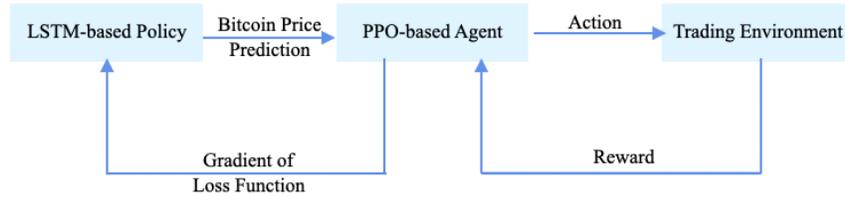

Fig. 3. Scheme of the Proposed Framework

## 4. Experiments

*4.1 Data Preparation*

In this study, the data set comes from the website cryptodatadownload[1]. There are 30984 valid records, covering the period from 4:00 a.m. on Aug 17$^{th}$, 2017 to 0:00 a.m. on Feb 27$^{th}$, 2021.

The bitcoin price fluctuates severely with obvious seasonality, that is, the time series changes with a trend as time goes by, and such internal trend affects the prediction. Hence, the difference method is applied to eliminate the trend. Specifically, two adjacent values are subtracted to get their variation. Only the processed data will be analyzed after difference. Thus, only changes between continuous data will be focused, ignoring the inner seasonality formed by the accumulation of the data itself. After prediction, the result will be restored by a reversed operation.

This study evaluates the stability of the processed series by the enhanced Dickey fuller test (ADF test), and the p value is 0.00, which verifies that the hypothesis of 0 can be rejected, equivalent to the fact that the time series after difference is stable.

The first 70% of the data set in chronological order as the training set, then the next 10% is set as the valid set and the rest as the test set. In order to enhance the speed of convergence, the data is normalized before being input into the model for training via the minimum-maximum value normalization method, whose definition is shown in (17):

$$origin_i^* = (y_{max} - y_{min}) \times \frac{origin_i - origin_{min}}{origin_{max} - origin_{min}} + y_{min} \quad (17)$$

where $origin_{min}$ and $origin_{max}$ represent the minimum and maximum values of the unprocessed data set respectively; $y$ denotes the normalized data set.

---

[1] https://www.cryptodatadownload.com/

*4.2 Policy Comparison*

Referring to previous related research, this study compares the performances of predicting static data. The structures of benchmarks are listed as follows:

- SVM: Adopt the package sklearn and set it as default without changing its structure and parameters.

- MLP: three levels, namely an input layer, an output layer, and a hidden layer.

- LSTM: four LSTM layers are set as the hidden layer to receive the input, whose activation function is ReLU; a dense network layer is set for the output; the activation function is linear, representing the linear relationship between the output of the upper node and the input of the lower node in the multi-layer neural network.

- TCN: TCN is an architecture based on convolutional network modeling of sequences. The structure and hyper-parameters of TCN used in this paper are consistent with those in the original paper [21].

- Transformer: Transformer is a seq2seq model with encoder and decoder. The encoder block is composed of 6 same layers combined by two sub-layers, namely a multi-head self-attention mechanism and a fully connected feed-forward network. There are residual connections and normalization in each sub-layer. The decoder is similar to the encoder, but has an additional attention sub-layer. This study adopts the original structure and hyper-parameters proposed in [22].

The selected models are built as the previous description. The loss function of all the above-mentioned approaches is Mean Square Error (MSE), which is also chosen as the evaluation index on the test data set.

The time step is opted as 10 to enables the sequence length to fit all selected approaches. Hyper-parameters for MLP-based and LSTM-based methods are decided through the grid method. The batch size is selected from {256, 128, 64, 32, 16}, the hidden unit number is selected from {200, 100, 50, 25} and the dropout rate is selected from {0.5, 0.4, 0.3, 0.2, 0.1}.The best performance's corresponding hyper-parameter combination for LSTM is {batch size = 32, hidden unit number = 50, dropout rate = 0.2}. The optimization is Adam, which can dynamically adjust the learning rate of each parameter. The initial learning rate is 0.001 referring to previous experience. Training epochs is 300, but "early stop" is set, that is, if MSE on the valid data set does not drop within 5 epochs, then the defined learning rate of Adam reduces it to 20% of its own; if MSE on the valid data set does not drop within 10 epochs, then the training process stops.

Table 1 reveals the performances in predicting bitcoin price of above-mentioned approaches.

Table. 1 Performances of Selected Approaches

| Approaches | Results of Static Prediction | | |
|---|---|---|---|
| | Best Test MSE | Training Time[1] / Epoch | Stopping Epoch |
| SVM | 0.0084 | 0.01s | 11 |
| MLP | 0.0251 | 1.72s | 30 |
| LSTM | 0.0015 | 13.25s | 24 |
| TCN | 0.01327 | 68.28s | 103 |
| Transformer | 0.0044 | 10.81 | 37 |

---

[1] CPU: 2 GHz 4-core Intel Core i5

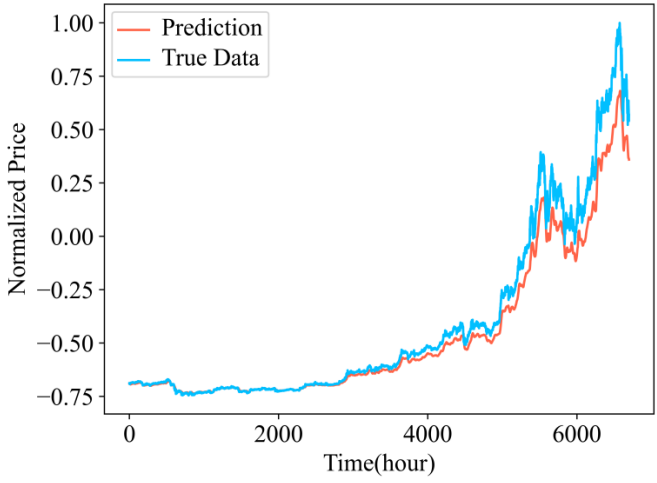

Fig. 4-1 SVM Prediction Results

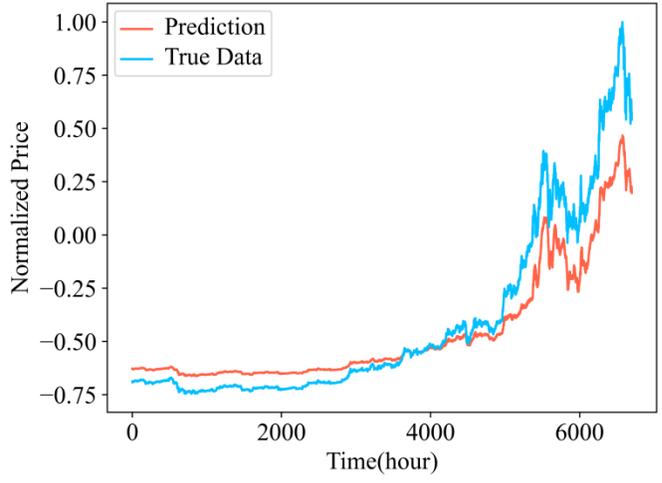

Fig. 4-2 MLP Prediction Results

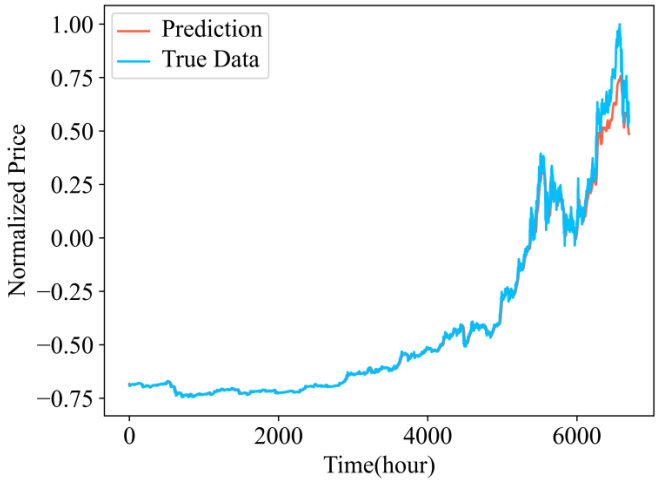

Fig. 4-3 LSTM Prediction Results

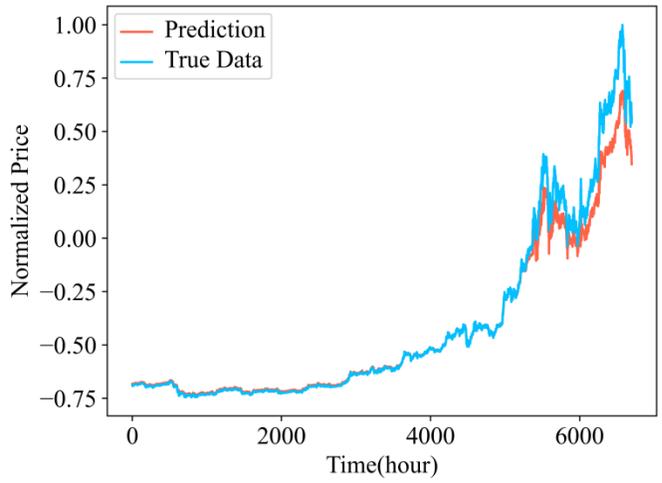

Fig. 4-4 TCN Prediction Results

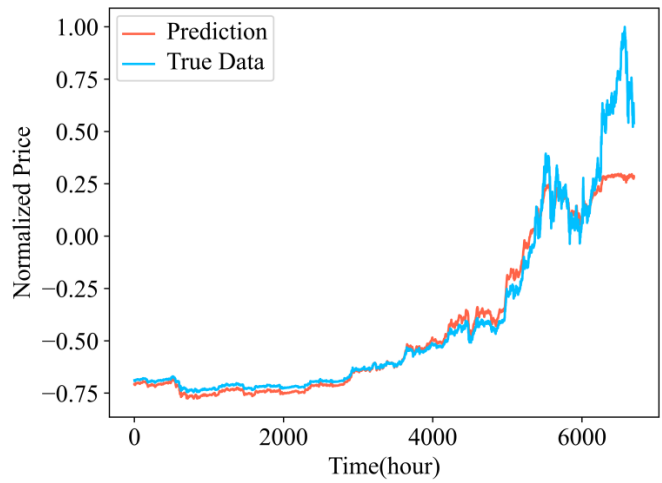

Fig. 4-5 Transformer Prediction Results

Fig. 4. Prediction Results of Selected Approaches

Fig. 4 shows the prediction results of selected approaches, in which the red curves denote the actual price and the blue curves denote the predicted price.

From the above experimental results, LSTM outperforms among the methods that have been proven to be effective in time-series prediction. LSTM is an excellent variant of RNN, which inherits the characteristics of most RNN models and solves the vanishing gradient problem caused by the gradual reduction of the gradient back-propagation process. Therefore, LSTM is extremely suitable for mining time dependences in long sequences and learning fitting patterns from historical information.

Although many studies have shown that TCN and Transformer are superior to LSTM. However, their major advantage is the ability to be parallelized when dealing with large-scale data. However, in this research scenario, the data scale of bitcoin's historical price is limited, and there is no need for parallel processing, so the advantages of TCN and Transformer cannot be brought into play.

On the other hand, the structure of transformer and TCN is more complex, with more internal parameters and higher requirements for the size of data involved in training, which may be one of the reasons why they do not perform as well as LSTM.

In a word, LSTM is an extraordinary choice for predicting the price of bitcoin, which is therefore selected to generate the policy function for the reinforcement learning agent.

## 4.3 Environmental Parameters of PPO-based Agent

To build a transaction agent according to the algorithm PPO described above, this study uses the Gym environment provided by OpenAI. The initial holding amount of the agent is U.S. $10,000; the handling fee required is 0.25% of each transaction amount; the maximum sliding point rate is 2%; the training frequency in each iteration is set as the length of the training set. The minimum trading unit is 0.125 bitcoin.

There are three possible actions each time (i.e. buy, sell and hold), so there are 24 actions in the action space. For the agent, 70% of the data set is split into the training set, 10% is the valid set, and the rest 20% is the test set. All comparisons of returns are based on the test set.

## 4.4 Benchmarks for Trading Strategies

Benchmarks are chosen from technical strategies, including the Buy and Hold, the Golden Cross/Death Cross strategy, the Momentum strategy [23], the Variable Moving Average (VMA) Oscillator-based strategy [24], and a non-named strategy defined by [12].

(1) **Buy and Hold**: Buys BTC at time $t=0$ with the initial capital and sell it only once at the time when profits are evaluated.

(2) **Golden Cross/Death Cross strategy**: (i)If at time $t$, the average increase from $t$-5 to $t$ is higher than the average increase at $t$-20 to $t$ by $r\%$ ($r > r_0$), meaning it achieves the golden cross, and then buy $r \times u$ bitcoins, where $r_0$ and $u$ are preset quantities. $r_0$ is usually defined as 5 and $u$ is generally 0.05; (ii) If at time $t$, the average decline from $t$-20 to $t$ is higher than the average decline from $t$-5 to $t$ by $r\%$ ($r > r_0$), meaning it achieves the death cross, then sell $r \times u$ bitcoins.

(3) **VMA Oscillator-based strategy**: first calculate the long and short period moving average, that is:

$$LongMA_{n,t} = \frac{1}{n}\sum_{t=0}^{n-1}\log(P_t) \; ; \; ShortMA_{n,t} = \log(P_t)$$

where $P_t$ is the price at time $t$. Then if $ShortMA_{n,t}$ is larger than $LongMA_{n,t}$, buy $u$ bitcoins. $u$ is also set as 0.25. $n$ is set as 50 according to [25].

(4) **Improved Momentum strategy**: According to the prediction results of LSTM, if the price at the time point $t+1$ is higher than that at $t$, $u$ bitcoins will be bought; otherwise, if the price at the time point $t+1$ is lower than that at $t$, $u$ bitcoins will be sold. $u$ is also set as 0.25.

(5) **Non-named strategy**: Take actions according to the *t*+1 prediction at the time point *t*. (i) Keep the cash and if the predicted price at *t*+1 is larger than that at *t*, then buy *u* bitcoins, otherwise, do nothing. Sell all bitcoins at the time when profits are evaluated. (ii) Buy as many bitcoins as it can. If the predicted price at *t*+1 is smaller than that at *t*, then sell *u* bitcoins, otherwise, do nothing. *u* is also set as 0.25.

The first utilizes no information while the second and the third strategies only learn from history. The fourth and the fifth methods take good advantage of prediction results obtained from the experiments in the previous section. Previous researchers tended to focus on complex asset allocation strategies or various technical indexes but lacked research on single-digit currency trading strategies. There is therefore an absence of widely recognized benchmarks. From this perspective, this study fills a gap in the field.

*4.5 Experiment Results of Trading*

The result of the proposed approach is shown in Fig. 5, where the green line denotes the tendency of net worth and the gray line denotes the tendency of bitcoin price.

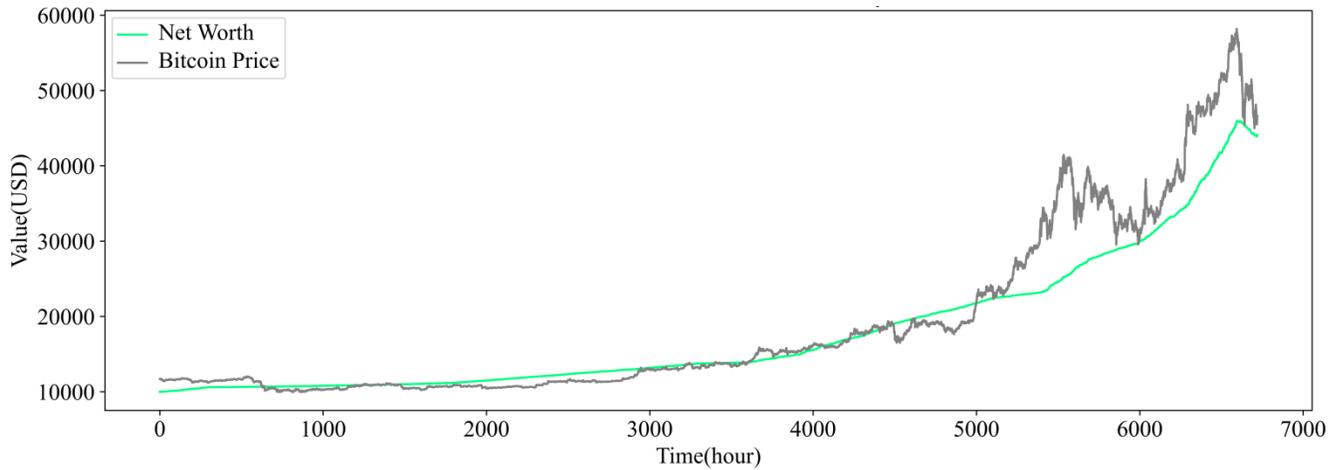

Fig. 5. Tendency for Net Worth of the Proposed Agent and Bitcoin Price

The proposed model outperforms significantly, whose profit rate on the test data set is 39.16% higher than that of the best benchmark. Details are shown in Table. 2.

Table. 2 Profit Rate of Various Strategies

| Strategies | Profit Rate / % |
|---|---|
| Proposed Framework | 341.28 |
| Buy and Hold Strategy | 301.79 |
| Golden Cross/Death Cross Strategy | 26.11 |
| VMA Oscillator-based Strategy | 302.12 |
| Improved Momentum Strategy | 8.17 |
| Non-named Strategy (i Buy) | 309.61 |
| Non-named Strategy (ii Sell) | - 0.84 |

Compared with the benchmarks, the proposed strategy has obvious advantages. Its profit rate is 39.49% higher than that of the Buy and Hold strategy, 39.16% higher than that of the VMA Oscillator-based strategy, 315.17% higher than that of the Golden Cross/Death Cross strategy, 303.06% higher than that of the Improved Momentum strategy, 31.67% higher than that of the Non-named strategy (i), and 342.12% higher than that of the Non-named strategy (ii). It is validated that the proposed approach can earn excess returns no matter whether the market rises or falls on a whole.

In the focused period, bitcoin experienced a round of surges, and its abnormal rise made a large number of high-frequency trading strategies fail. The main reason is that high-frequency trading strategies usually focus on short-term earnings, ignoring the long-term price trend. Although the agent constructed in this paper is based on LSTM forecasting, it still learns the long-term market trend information in the process of interacting with the environment and demonstrates its benign performance and potential in the subsequent high-frequency trading.

In the future, when the price of bitcoin tends to be flat (i.e. fluctuates around its real value), high-frequency trading is more likely to obtain excess returns. At that time, the high-frequency transaction agent constructed in this paper will play an important role. On the other hand, only the proposed method considers risk by introducing Omega Ratio as the reward signal, indicating that it has a more practical value in real transactions.

The proposed model succeeds to achieve profitability, though it still has space for improvement. The agent may not be able to allocate assets evenly, resulting in excessive purchases of bitcoin in the early stage, leading to the lack of cash after that. Therefore, it may not be able to seize the opportunity to increase positions.

## 5. Conclusion

This study demonstrates the prediction accuracy on the bitcoin price of different kinds of advanced machine learning-based approaches. The proposed framework makes full use of LSTM and PPO to build an agent that can automatically generate bitcoin trading strategies and earns excess profit based on DRL.

First, it compares mainstream sequence-prediction techniques including SVM, MLP, LSTM, TCN, and Transformer to find a model best fitting the bitcoin price time-series. The experimental shows that LSTM outperforms and this paper explains its outstanding performance in this application scenario. Thus, LSTM can fit the bitcoin price best and play the role of constructing the policy function.

Second, it proposes a general framework for automatically generating a high-frequency trading strategy with a PPO-based agent. It extends the application of DRL from action games to trading. Moreover, the agent is proved to have better performances than traditionally popular benchmarks and be able to earn high excess returns. The reason for different profit rates obtained from different methods is discussed and it is analyzed why the proposed framework outperforms in the bitcoin's shock and surge.

Third, this study opens the door to research on building a single cryptocurrency high-frequency trading strategy, making it possible to automatically make decisions in the rapidly changing digital currency market. What's exciting is that the proposed framework can be easily extended to any tradable financial asset theoretically, such as stocks, features, and other kinds of cryptocurrency (e.g., Ethereum (ETH) and Dogecoin). Since this approach only involves digital information rather than ad-hoc knowledge, the price time series of all these assets can be directly inputted into the framework to automatically generate a trading signal, though more experiments are needed to explore the performances.

To sum up, this paper not only contributes to exhibit the wide application of this cutting-edge algorithm outside the action game field but also complements traditional high-frequency quantitative trading methods. It fills a gap in the field of single-assent transaction strategy construction leveraging state-of-the-art techniques only with information of history price. Additionally, it displays potential factors in the trading simulation, which will enlighten researchers and investors.

Last but not least, there is still room for exploration. First, the reward strategy of this paper is relatively simple. A more realistic strategy is to introduce more technical indicators to enrich the interaction between agent and environment, and then adopt automated reinforcement learning to automatically select a technical indicator suitable for the input data as the reward signal [26], which undoubtedly simplifies the construction of a predictive model. Second, the agent with trained weights might provide reference to investing other kinds of cryptocurrency by adopting the idea of transfer learning and fine-tuning. Third, it would be interesting to develop a privacy-preserving bitcoin transaction strategy through motivating bitcoin owners to participate in federated learning [27]. Finally, combined with manual work, the proposed method may achieve a more controllable risk investment strategy in practice.

## Acknowledgements

This work is partly supported by the Natural Science Foundation of Jilin Province, China under Grant No. YDZJ202101ZYTS149.